\documentclass[a4paper,conference]{IEEEtran}

\usepackage{subfigure}
\usepackage{comment}
\usepackage{wrapfig}
\usepackage{algorithm}
\usepackage{algorithmic}
\usepackage{amsmath,amssymb}
\usepackage{graphicx}
\usepackage{url}
\usepackage{color}
\usepackage{times}
\usepackage{epsfig}
\usepackage{tabularx}
\usepackage[pagebackref=true,breaklinks=true,colorlinks,bookmarks=false]{hyperref}

\DeclareMathOperator*{\argmax}{arg\,max}

\usepackage{etoolbox}
\makeatletter
\patchcmd{\@makecaption}
  {\scshape}
  {}
  {}
  {}
\makeatother

\begin{document}

\title{The Sound of Bounding-Boxes}

%
	
\author{\IEEEauthorblockN{Takashi Oya\IEEEauthorrefmark{1},
Shohei Iwase\IEEEauthorrefmark{1},
Shigeo Morishima}
\IEEEauthorblockA{Waseda Research Institute for Science and Engineering, Japan, 513-1 Tsurumaki Waseda Shinjuku \\Email: oya\_takashi@ruri.waseda.jp, sh.iwase@fuji.waseda.jp, shigeo@waseda.jp}}


\maketitle

\begin{abstract}
In the task of audio-visual sound source separation, which leverages visual information for sound source separation, identifying objects in an image is a crucial step prior to separating the sound source. 
However, existing methods that assign sound on detected bounding boxes suffer from a problem that their approach heavily relies on pre-trained object detectors. 
Specifically, when using these existing methods, it is required to predetermine all the possible categories of objects that can produce sound and use an object detector applicable to all such categories. 
To tackle this problem, we propose a fully unsupervised method that learns to detect objects in an image and separate sound source simultaneously. 
As our method does not rely on any pre-trained detector, our method is applicable to arbitrary categories without any additional annotation.
Furthermore, although being fully unsupervised, we found that our method performs comparably in separation accuracy.
\end{abstract}


\section{Introduction}
In the world around us numerous sounds exist at the same time such as the sound of people speaking, the engine sound of cars going by, and the roar of planes flying overhead. Humans have the capability to identify and isolate individual sounds in such environments, and this is particularly true when we are able to see the objects emitting the sound. This task of separating sounds given visual guidance but in an automatic manner is called audio-visual sound source separation and has been studied in the field of computer vision and signal processing. 
More specifically, in this task, given an image with multiple sound-producing objects and the corresponding monaural audio as input, the objective is to locate each sound-producing object in the image and to separate the monaural audio into individual sounds associated with the objects.

One promising direction for audio-visual sound source separation is the usage of pre-trained object detectors to identify sound-producing objects prior to separating the sound source. However, in this case, an abundant amount of annotated data is required to train the detector, and moreover, categories whose annotations are not available can not be handled. 

\begin{figure}
    \centering
    \scalebox{1}{
        \includegraphics[width=\linewidth]{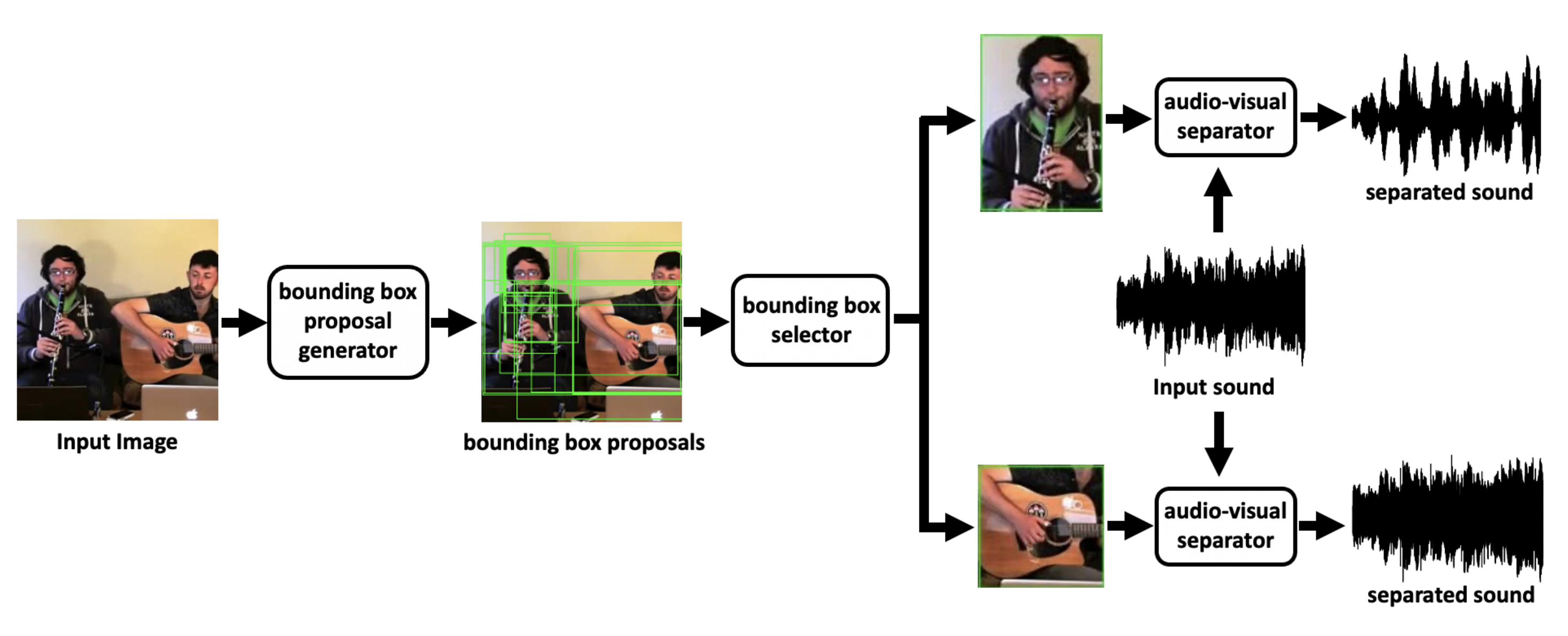}
    }
    \caption{
Overview of our system.
The system is composed of 3 modules, the bounding box proposal generator, the bounding box selector, and the audio-visual separator. First, the bounding box proposal generator gives a proposal of bounding boxes by using a conventional algorithm called EdgeBoxes \cite{edgeboxes}, which is based on the edge information of an image. Then, the bounding box selector selects the bounding boxes which correspond to the regions that are producing sound. Finally, the audio-visual separator returns the separated sound for the selected bounding boxes. In the example above, the bounding box that contains a clarinet and the one that contains an acoustic guitar is selected by the bounding box selector. Then, the audio-visual separator returns the sound of the clarinet and the sound of the acoustic guitar by extracting the corresponding components from the input sound.
}
    \label{fig:overview}
\end{figure}

To tackle this problem, we propose a fully unsupervised method that learns to detect objects in an image and separate sound source simultaneously. Unlike existing methods, our method can be trained in an end-to-end manner including the object detection process. 
As shown in Fig. \ref{fig:overview}, our proposed method is composed of 3 modules: the bounding box proposal generator, the bounding box selector, and the audio-visual separator. 
The bounding box proposal generator takes an image as input and returns bounding box proposals. For this module, we leverage EdgeBoxes \cite{edgeboxes} which has been used in conventional object detection models (e.g. R-CNN \cite{R_CNN}).
The bounding box selector is a novel module, which takes the bounding box proposals as input and returns the indices of the selected bounding boxes. Intuitively, this module learns to select bounding boxes that are likely to produce sound. 
The audio-visual separator takes the image features of the selected bounding boxes and the sound source as inputs and returns the separated sound for each bounding box.
To make the overall architecture differentiable and trainable, we used a categorical reparameterization trick called Straight-Through Gumbel Softmax \cite{gumbel} for the bounding box selector.  
Owing to this technique, our model can be trained end-to-end including the object detector.
In contrast to the pre-trained detector based method which fundamentally can not handle categories whose annotations are not available, by design our model is applicable to such categories.
In our experiments, we compared separation accuracy with the pre-trained detector based method, and found that our proposed model performs comparably when validated on known categories for the pre-trained detector.
\section{Related Work}
In this section, we will go over previous studies on audio-visual sound source separation, and sound source localization which is a closely related task.
\\
\\
\noindent
\textbf{Sound source localization}
Sound source localization is a traditional problem in robotics and signal processing that uses audio information from multiple microphones to determine the direction and position of a sound source. In the field of computer vision, sound source localization is regarded as the prediction of the location of sound sources in an image and is also sometimes called audio-visual sound source localization.
There have been many sound source localization methods in the field of computer vision, e.g. mutual information and CCA \cite {synchrony, pixels_that_sound}, CAM-based \cite {multisensory, cooperative_learning}, attention mechanism based \cite {objects_that_sound, learning_to_localize, localize_hard_way, potential_localize}, those that utilize motion information \cite {event_localization, sound_of_motions, gesture}. Sound source localization and audio-visual sound source separation are closely related because it is necessary to identify the position of the sound source in an image in order to perform audio-visual sound source separation. Inevitably, sound source localization must be performed prior to audio-visual source separation.
\\
\\
\noindent
\textbf{Audio-visual sound source separation.} 
Audio-visual sound source separation is the task of separating sound sources using visual information. There are many methods for audio-visual sound source separation, e.g. NMF \cite{motion_informed, two_multimodal, learning_to_separate}, subspace method \cite{independent_components, sparsity}, mix-and-separate method \cite{sound_of_pixels, multisensory, co_segmentation, co_separating, sound_of_motions, gesture, visualvoice, visual_graph}, use of facial information \cite{cocktail, visual_speech_enhancement, conversation_enhancement, blind_audiovisual_source_separation}, use of facial information without lip movements \cite{visualvoice}, use of the visual structure of the scene as graph \cite{visual_graph}. 
In audio-visual sound source separation, it is essential to identify the location of the sound sources in advance, and therefore several existing methods identify the location of the sound source using bounding boxes via pre-trained detectors \cite{co_separating, cocktail}. For example, Gao \& Grauman \cite{co_separating} train an object detection model using the samples in the instrument categories of the Open Images dataset \cite{openimages}. However, in order to adapt the model to new categories, it is necessary to prepare new annotations for these additional categories and retrain the object detection model. 

Alternatively, there are methods that do not assign sound to bounding boxes but instead assign sounds to pixels \cite{sound_of_motions, sound_of_pixels}. These methods are trained in an unsupervised manner and do not require annotations.
One drawback of these methods is that the locations of the sound sources are not explicitly identified.  Also, when the sound of each sound source is desired, the pixel that best represents the actual sound is ambiguous. In contrast to these methods, our method can explicitly identify the location of the sound sources using bounding boxes.

\begin{figure*}[h]
    \centering
    \scalebox{0.67}{\includegraphics[width=\linewidth]{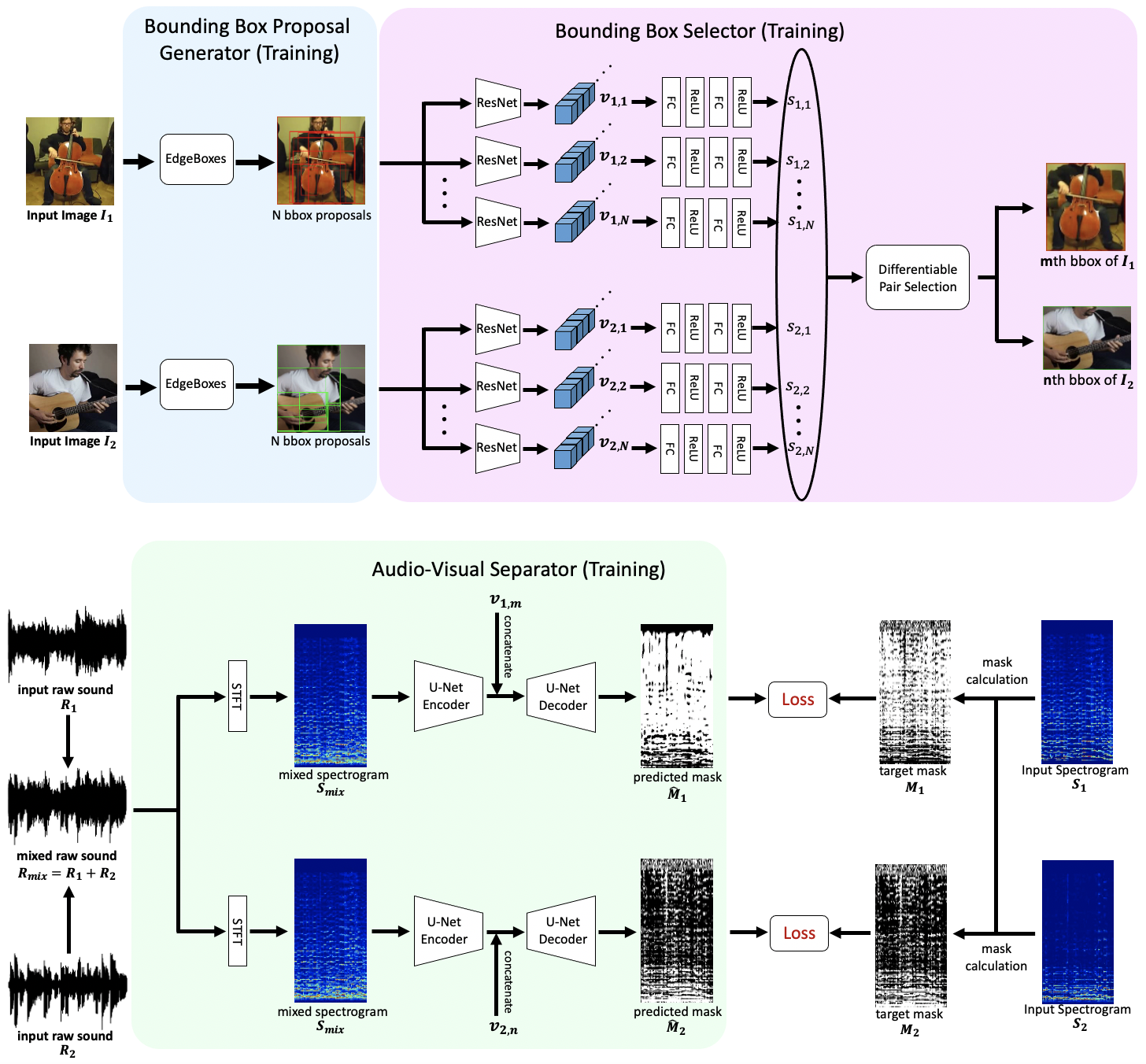}}
    \caption{
    The detailed network architecture in training phase. In training phase, a pair of image/sound that is randomly sampled from the training set is used as input. The bounding box selector selects the bounding box that is most likely to produce sound from the bounding box proposals of each image in a differentiable manner. 
    Then, sound inputs are artificially mixed and the audio-visual separator learns to separate the mixed sound according to the selected bounding boxes. The predicted mask indicates which pixels of the mixed spectrogram correspond to each of the selected bounding boxes.
    }
    \label{fig:arch1}
\end{figure*}
\begin{figure*}[h]
    \centering
    \scalebox{0.67}{\includegraphics[width=\linewidth]{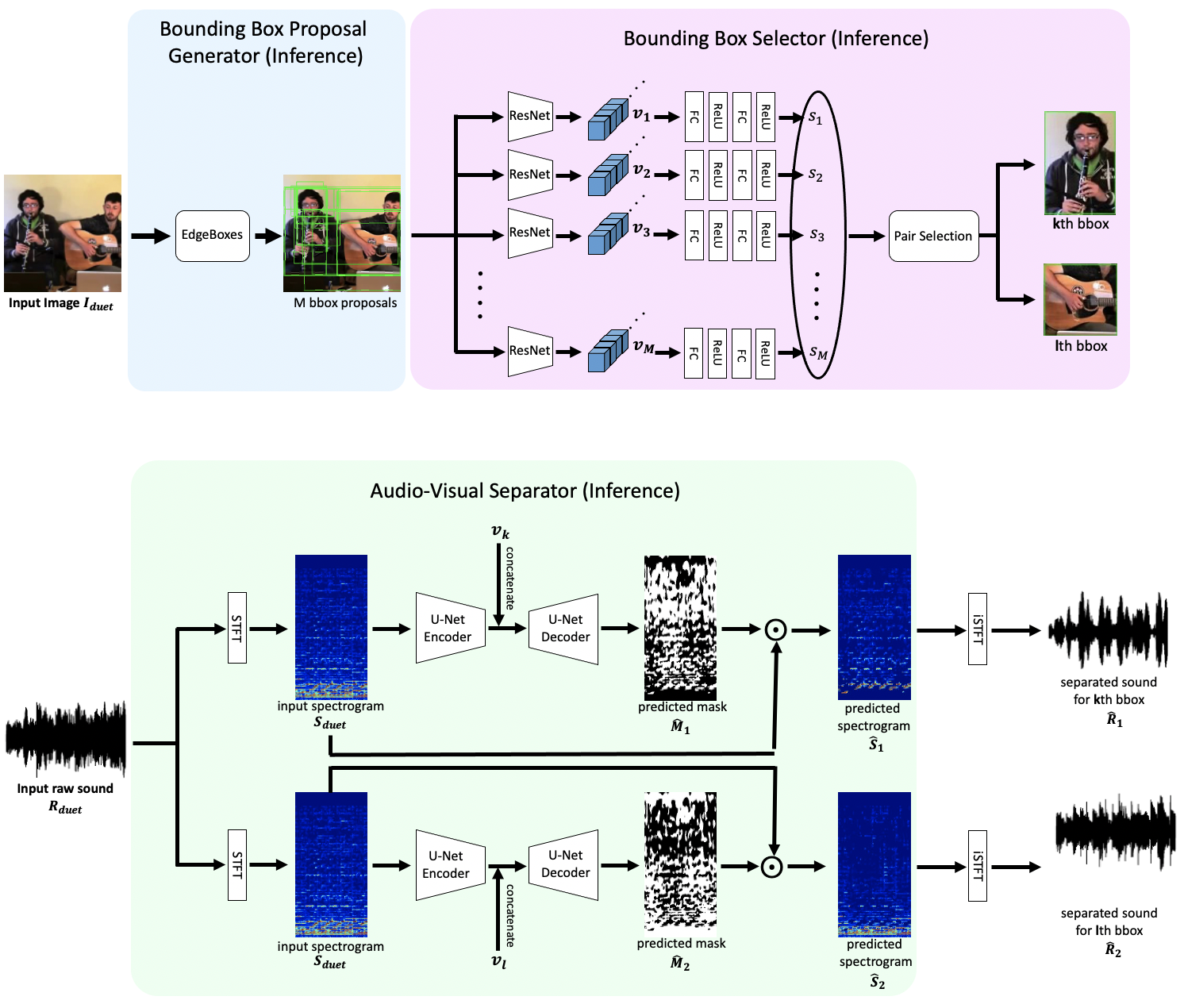}}
    \caption{
    The detailed network architecture in inference phase. In inference phase, an image and sound containing multiple sound sources are used as inputs. The bounding box proposal selector selects the pair of bounding boxes that are most likely to produce sound. Then, the audio-visual separator separates the input sound and returns the sound corresponding to each of the selected bounding boxes. 
    }
    \label{fig:arch2}
\end{figure*}
\section{Proposed Framework}
Following existing studies on audio-visual sound source separation, we adopt different procedures for the training phase and the inference phase. In the training phase, a pair of videos that each contains a single sound-producing object is used as input, e.g. a video of a guitar being played and a video of a violin being played. The accompanying monaural audio of each video are mixed into a single audio to artificially create an audio environment containing multiple objects. The task at hand during training is to locate the sound-producing object in each video via bounding boxes, and to separate the mixed audio into individual audio that matches the located objects, i.e. reconstruct the original single audio. The detailed process of the training phase is shown in Fig. \ref{fig:arch1}. In the inference phase, the difference is that the input is a single video containing multiple sound-producing objects. Therefore, multiple sound-producing objects are detected in the single input video, unlike the training phase where only a single object can be found in each video. The detailed process of the inference phase is shown in Fig. \ref{fig:arch2}. The rationale for adopting different procedures for training and inference is that, for videos with multiple sound-producing objects, obtaining the ground-truth audio for each object is impractical.

\subsection{Bounding Box Proposal Generator}
The bounding box proposal generator is a module that takes an image as input and returns a set of possible bounding boxes. For this module, we simply employed EdgeBoxes \cite{edgeboxes}, a conventional algorithm that has been used for training object detection models.
In training phase, N bounding boxes are obtained for both images ($I_1$, $I_2$). In inference phase, M bounding boxes are obtained for the duet image ($I_\text{duet}$), i.e. an image containing multiple sound-producing objects. In both phases, the images are cropped using these bounding boxes and passed on to the bounding box selector.  

\subsection{Bounding Box Selector} 

\noindent
\textbf{Training phase}
In training phase, the bounding box selector takes 2N cropped images. First, DilatedResNet18 \cite{dilatedresnet} is applied to the input images, and the image features $\boldsymbol{v_{1, i}} \in R^C \; (1 \leq i \leq N)$ are obtained from the first image and the image features $\boldsymbol{v_{2, j}} \in R^C \; (1 \leq j \leq N)$ are obtained from the second image. Here, $C$ denotes the dimension of the image features. Then, positive scalar values $s_{1, i}, s_{2, j} \in R       $ are obtained from each image features, by applying fully-connected layers (FC) as shown below:
\begin{align}
    s_{1, i} = {\rm ReLU}({\rm FC}({\rm ReLU}({\rm FC}(\boldsymbol{v_{1, i}})))
\label{fc1}
\end{align}
\begin{align}
    s_{2, j} = {\rm ReLU}({\rm FC}({\rm ReLU}({\rm FC}(\boldsymbol{v_{2, j}}))).
\label{fc2}
\end{align}
Intuitively, $s_{1,}$ and $s_{2, j}$ indicate how likely it is that sound is being produced from the region of the corresponding bounding box.
After that, the pair of bounding boxes that are most likely
to produce sound can be selected by the following equation: 
\begin{align}
\label{train_pair_select}
    (m, n) = \argmax_{i, j} s_{1, i} s_{2, j}.
\end{align}
However, Eq. \ref{train_pair_select} itself is not differentiable, so it can not be used in training. To solve this problem, we made the process differentiable by introducing a categorical reparameterization trick, which will be explained in the next section.  
\\
\\
\noindent
\textbf{Differentiable Pair Selection}
To make the process in Eq. \ref{train_pair_select} differentiable, We first calculated 
\begin{align}
P_{i, j} = \frac{s_{1, i} s_{2, j}}{\sum_{i, j} s_{1, i} s_{2, j}}.
\end{align}
Then, a categorical reparameterization trick called Straight-Through (ST) Gumbel Softmax \cite{gumbel} is leveraged to make $P_{i, j}$ discrete.

\begin{align}
D_{i, j} = \text{ST\_Gumbel\_Softmax}(P_{i, j})
\end{align}
Finally, the image features corresponding to the selected bounding boxes ($\boldsymbol{v_{1, m}}$, $\boldsymbol{v_{2, n}}$) can be obtained by the following equation, in a differentiable manner. 
\begin{align}
\boldsymbol{v_{1, m}} = \sum_{i} (\boldsymbol{v_{1, i}} \sum_{j} D_{i, j})
\end{align}

\begin{align}
\boldsymbol{v_{2, n}} = \sum_{j} (\boldsymbol{v_{2, j}} \sum_{i} D_{i, j}).
\end{align}
\\
\\
\noindent
\textbf{Inference phase}
In inference phase, the bounding box selector takes M cropped images. Same as the training phase, DilatedResNet18 is applied to the input images and the image features $\boldsymbol{v_k} \in R^C \; (1 \leq k \leq M)$ are obtained. Then, positive scalar values $s_{k} \in R         \; (1 \leq k \leq M)$ are obtained by 
\begin{align}
    s_{k} = {\rm ReLU}({\rm FC}({\rm ReLU}({\rm FC}(\boldsymbol{v_{k}})))
\label{fc3}
\end{align}
After that, the pair of non-overlapping bounding boxes that  is most likely to produce sound is selected by
\begin{align}
\label{test_pair_select}
    (k, l) = \argmax_{(i, j) \in \text{NO}, i < j} s_{i} s_{j}.
\end{align}
Here, $\text{NO}$ is decided by the following condition.
\begin{align}
\label{NO}
 \text{NO} = \{(i, j) | i \text{th bbox and } j \text{th bbox do not overlap}\}.
\end{align}
Finally, the image features corresponding to the selected bounding boxes ($\boldsymbol{v_{k}}$, $\boldsymbol{v_{l}}$) are used as the input of the audio-visual separator. 

\subsection{Audio-Visual Separator}

\noindent
\textbf{Training phase}
In training phase, the audio-visual separator takes an artificially mixed raw sound $R_\text{mix}$ and the image features ($\boldsymbol{v_{1, m}}$, $\boldsymbol{v_{2, n}}$) as inputs. Here, $R_\text{mix}$ can be calculated by $R_\text{mix} = R_1 + R_2$, where $R_1$ and $R_2$ denote an input raw sound pair. First, Short-Time Fourier Transform (STFT) is applied to $R_\text{mix}$ and the spectrogram $S_\text{mix}$ is obtained. Then, the spectrogram $S_\text{mix}$ is used as the input of the U-Net and each of the image features ($\boldsymbol{v_{1, m}}$, $\boldsymbol{v_{2, n}}$) are concatenated with the intermediate features of the U-Net. The output of the U-Net is the spectrogram masks ($\widehat{M}_1$, $\widehat{M}_2$) corresponding to each of the image features. Finally, per-pixel cross entropy loss is calculated between the predicted masks ($\widehat{M}_1$, $\widehat{M}_2$) and the target masks ($M_1$, $M_2$), where $M_1$ and $M_2$ are calculated by 
\begin{align}
  M_1(x, y) = \begin{cases}
    1 & (\frac{S_1(x, y)}{S_1(x, y) + S_2(x, y)} > 0.5) \\
    0 & (otherwise)
  \end{cases}
\end{align}
\begin{align}
  M_2(x, y) = \begin{cases}
    1 & (\frac{S_2(x, y)}{S_1(x, y) + S_2(x, y)} > 0.5) \\
    0 & (otherwise)
  \end{cases}.
\end{align}
These masks are called ``binary mask'' in existing studies, and are empirically known to work well. The per-pixel cross entropy loss is the only loss used to train the model.
\\
\\
\noindent
\textbf{Inference phase}
In inference phase, the audio-visual separator takes a raw sound $R_\text{duet}$ and the image features ($\boldsymbol{v_k}$, $\boldsymbol{v_l}$) as inputs. First, same as training phase, STFT is applied to $R_\text{duet}$, and the spectrogram $S_\text{duet}$ is obtained. Then, the spectrogram $S_\text{duet}$ is used as the input of the U-Net, and each of the image features ($\boldsymbol{v_k}$, $\boldsymbol{v_l}$) are concatenated with the intermediate features of the U-Net. The output of the U-Net is the spectrogram masks ($\widehat{M}_1$, $\widehat{M}_2$) corresponding to each of the image features. These masks are multiplied with the input spectrogram, and then the predicted spectrograms for both bounding boxes ($\widehat{S}_1$, $\widehat{S}_2$) are obtained. Finally, inverse STFT is applied to the predicted spectrograms and the separated raw sound ($\widehat{R}_1$, $\widehat{R}_2$) is obtained.  

\subsection{Constraint on predicted masks using softmax function}
\label{softmax}
We found our method can be easily improved by introducing a simple constraint that forces the sum of the predicted masks to be equal to 1. Specifically, we replaced the sigmoid layer, which is used as the final activation function in the U-Net, with a softmax layer. When using the sigmoid layer, the masks are calculated by
\begin{align}
    \widehat{M}_1 = \frac{1}{1 + \exp(-U_1)}, \widehat{M}_2 = \frac{1}{1 + \exp(-U_2)} 
\end{align}
, where $U_1$ and $U_2$ are the outputs of the U-Net decoder prior to the last activation function.
On the other hand, when using the softmax layer, the masks are calculated as follows. 
\begin{align}
    \widehat{M}_1 = \frac{\exp(U_1)}{\exp(U_1) + \exp(U_2)}
\end{align}
\begin{align}
    \widehat{M}_2 = \frac{\exp(U_2)}{\exp(U_1) + \exp(U_2)}.
\end{align}
Because of the softmax function, it is obvious that the sum of the predicted masks is equal to 1. It should be noted that the sum of the separated sound is always equal to the original sound in such a situation.

\subsection{Hyperparameters}
For the hyperparameters of audio-visual separator and DilatedResNet18, we followed the official implementation of Pixelplayer \cite{sound_of_pixels}, except that the number of the frames in video is 1. For the hyperparameters of bounding box selector, we used $N = 10$, $M = 80$, C = 32. The number of the hidden layers in FCs of the bounding box selector is 128.

\section{Experiments}
\label{sec:experiments}
\begin{figure*}[h]
    \scalebox{0.70}{
        \begin{tabular}{p{7.6em}p{7.6em}p{7.6em}p{0.04em}|p{7.6em}p{7.6em}p{0.04em}|p{7.6em}p{7.6em}}
            \hfil \hfil 
            & \multicolumn{2}{c}{\qquad Example pair 1}
            & \hfil \hfil
            & \multicolumn{2}{c}{\qquad Example pair 2}
            & \hfil \hfil
            & \multicolumn{2}{c}{\qquad Example pair 3}
            	\\
            	\vspace{0.1mm}
            	
            \begin{minipage}{7.6em}
                \centering
              Input images
            \end{minipage} &
            \begin{minipage}{7.6em}
                \centering
                \scalebox{0.380}
                {\includegraphics{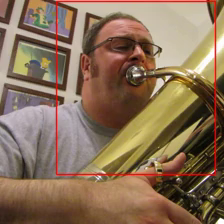}}
            \end{minipage} &
            \begin{minipage}{7.6em}
                \centering
                \scalebox{0.380}
                {\includegraphics{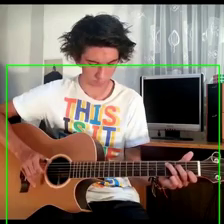}}
            \end{minipage} &
            \begin{minipage}{0.04em}
            \end{minipage} &
            
            \begin{minipage}{7.6em}
                \centering
                \scalebox{0.380}
                {\includegraphics{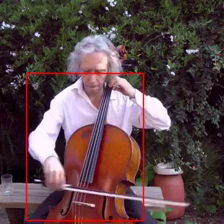}}
            \end{minipage} &
            
            \begin{minipage}{7.6em}
                \centering
                \scalebox{0.380}
                {\includegraphics{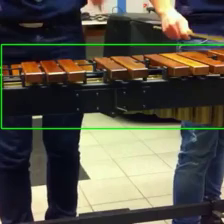}}
            \end{minipage} &
            \begin{minipage}{0.04em}
            \end{minipage} &
            
            \begin{minipage}{7.6em}
                \centering
                \scalebox{0.380}
                {\includegraphics{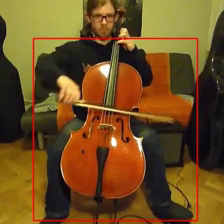}}
            \end{minipage} &
            \begin{minipage}{7.6em}
                \centering
                \scalebox{0.380}
                {\includegraphics{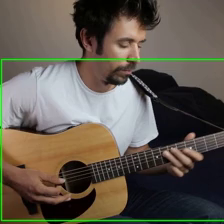}}
            \end{minipage} \vspace{0.001mm} \\
            \begin{minipage}{7.6em}
                \centering
                mixed spectrogram
            \end{minipage} &
            \multicolumn{2}{c}{ \quad \begin{minipage}{7.6em}
                \centering
                \scalebox{0.380}
                {\includegraphics{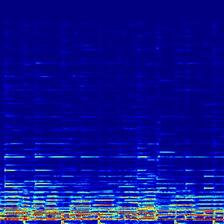}}
            \end{minipage}} &
            \begin{minipage}{0.04em}
            \end{minipage} &
            \multicolumn{2}{c}{ \quad \begin{minipage}{7.6em}
                \centering
                \scalebox{0.380}
                {\includegraphics{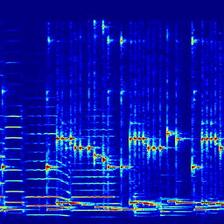}}
            \end{minipage}} &
            \begin{minipage}{0.04em}
            \end{minipage} &
            \multicolumn{2}{c}{ \quad \begin{minipage}{7.6em}
                \centering
                \scalebox{0.380}
                {\includegraphics{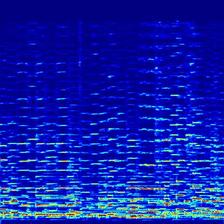}}
            \end{minipage}} \vspace{2mm} \\
            \begin{minipage}{7.6em}
                \centering
              Predicted spectrogram
            \end{minipage} &
            \begin{minipage}{7.6em}
                \centering
                \scalebox{0.380}
                {\includegraphics{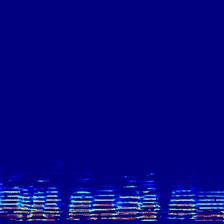}}
            \end{minipage} &
            \begin{minipage}{7.6em}
                \centering
                \scalebox{0.380}
                {\includegraphics{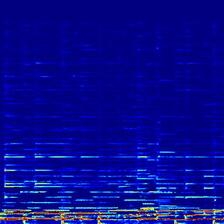}}
            \end{minipage} &
            \begin{minipage}{0.04em}
            \end{minipage} &
            
            \begin{minipage}{7.6em}
                \centering
                \scalebox{0.380}
                {\includegraphics{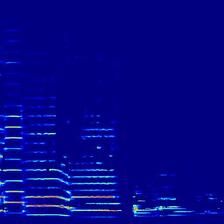}}
            \end{minipage} &
            
            \begin{minipage}{7.6em}
                \centering
                \scalebox{0.380}
                {\includegraphics{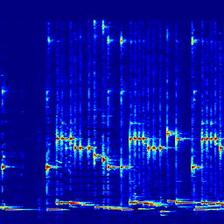}}
            \end{minipage} &
            \begin{minipage}{0.04em}
            \end{minipage} &
            
            \begin{minipage}{7.6em}
                \centering
                \scalebox{0.380}
                {\includegraphics{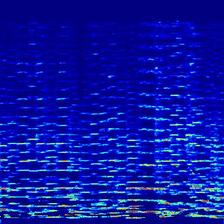}}
            \end{minipage} &
            \begin{minipage}{7.6em}
                \centering
                \scalebox{0.380}
                {\includegraphics{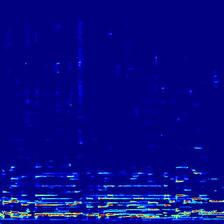}}
            \end{minipage} \vspace{0.001mm} \\
            \begin{minipage}{7.6em}
                \centering
                Ground truth spectrogram
            \end{minipage} &
            \begin{minipage}{7.6em}
                \centering
                \scalebox{0.380}
                {\includegraphics{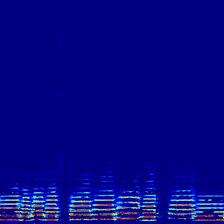}}
            \end{minipage} &
            \begin{minipage}{7.6em}
                \centering
                \scalebox{0.380}
                {\includegraphics{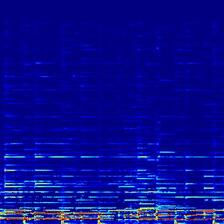}}
            \end{minipage} &
            \begin{minipage}{0.04em}
            \end{minipage} &
            
            \begin{minipage}{7.6em}
                \centering
                \scalebox{0.380}
                {\includegraphics{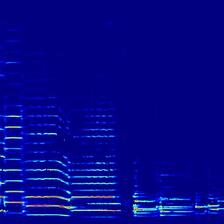}}
            \end{minipage} &
            
            \begin{minipage}{7.6em}
                \centering
                \scalebox{0.380}
                {\includegraphics{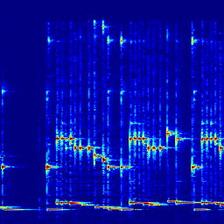}}
            \end{minipage} &
            \begin{minipage}{0.04em}
            \end{minipage} &
            
            \begin{minipage}{7.6em}
                \centering
                \scalebox{0.380}
                {\includegraphics{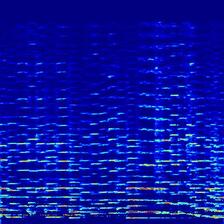}}
            \end{minipage} &
            \begin{minipage}{7.6em}
                \centering
                \scalebox{0.380}
                {\includegraphics{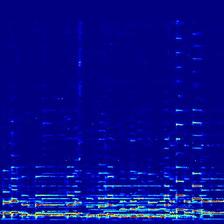}}
            \end{minipage} \vspace{0.001mm} \\

        \end{tabular}     }
    	\caption{Qualitative evaluation of our model. The model is trained using all categories and results for samples from the validation set are shown. It should be noted that the category of the left image in example pair 1 is `tuba', and the category of the right image in example pair 2 is `xylophone'. These categories are not included in the 7 single instrument categories that are available in the Open Images V4 dataset.}
    \label{fig:qualitative_solo} 
\end{figure*}
\subsection{Dataset and Data Preprocessing} 
For the training and evaluation, we collected 655 videos from the MUSIC dataset's \cite{sound_of_pixels} YouTube video IDs. The MUSIC dataset contains 20 categories, where 11 categories are of a single instrument (solo), and 9 categories are of two instruments (duet). We used 408 solo videos for training, 102 solo videos for validation, and 145 duet videos for testing. We split train/validation so that the proportion of the number of samples in each category is the same. The validation set is used to evaluate the separation accuracy, as the test set samples do not have ground truth sound of individual objects. The test set is used for the qualitative evaluation.

\subsection{Configurations}
To evaluate the separation accuracy of our model, we make a comparison with Co-separation \cite{co_separating} which is an audio-visual sound source separation model that locates sound-producing objects using bounding boxes in a supervised manner. In addition, we made a comparison with Pixelplayer \cite{sound_of_pixels}, which assigns sound to every pixel in the given image and is trained in an unsupervised manner. Although Pixelplayer does not explicitly identify the location of the sound-producing object and hence is tackling a slightly different problem, a comparison in separation accuracy can be made in the validation setting.
Because Co-separation depends on an object detector pre-trained on the Open Images V4 Dataset \cite{openimages}, comparison with Co-separation can only be made on the categories included in this dataset. Specifically, we used 7 out of 11 single instrument categories in the MUSIC dataset. The excluded categories are `clarinet', `erhu', `tuba', and `xylophone'. On the other hand, a comparison with Pixelplayer can be performed using all categories of the MUSIC dataset.  
Furthermore, we experimented both with and without the softmax constraint for our model and Co-separation. However for Pixelplayer, since the softmax constraint can not be applied in inference phase as sounds are assigned to every pixel in the image, the softmax constraint is not tested. When excluding the softmax constraint, the final activation function is simply a sigmoid.
For all models, the same loss function is used to ensure a fair comparison. 
\label{config}

\subsection{Evaluation}
For evaluation we employed SDR (Source to Distortion Ratio), SIR (Source to Interference Ratio) and SAR (Source to Artifacts Ratio).
It should be noted that SDR is considered as the overall metric that can evaluate the sound separation quality.

\subsection{Results and Analysis}
\noindent
\textbf{Quantitative evaluation} 
\begin{table}[t]
    \caption{Separation accuracy evaluated using only the 7 single instrument categories that are available in both the Open Images V4 Dataset \cite{openimages} and the MUSIC dataset \cite{sound_of_pixels}.}
    \centering
    \scalebox{1.0}{
\begin{tabular}{c|ccc}
\multicolumn{1}{l|}{}   & SDR                      & SIR                       & SAR                       \\ \hline
Ours (sigmoid)          & 7.78                     & 12.78                     & 11.52                     \\
Ours (softmax)          & \multicolumn{1}{l}{8.40} & \multicolumn{1}{l}{13.39} & \multicolumn{1}{l}{11.99} \\
Co-Separation (sigmoid) & 8.00                     & 13.14                     & 11.69                     \\
Co-separation (softmax) & 8.59                     & 13.77                     & 11.94                     \\
PixelPlayer (sigmoid)   & 7.76                     & 12.96                     & 11.07                     \\ \hline
\end{tabular}    }
    \label{tab:limited}
\end{table}
\begin{table}[t]
    \caption{Separation accuracy evaluated using all 11 single instrument categories in the MUSIC dataset \cite{sound_of_pixels}.}
    \centering
    \scalebox{1.0}{
\begin{tabular}{c|ccc}
                      & SDR  & SIR   & SAR   \\ \hline
Ours (sigmoid)        & 8.77 & 15.18 & 11.83 \\
Ours (softmax)        & 9.62 & 16.38 & 12.26 \\
PixelPlayer (sigmoid) & 8.41 & 15.81 & 10.92 \\ \hline
\end{tabular}    }
    \label{tab:all}
\end{table}
We reported SDR/SIR/SAR of our model, Pixelplayer \cite{sound_of_pixels} and Co-separation \cite{co_separating}, in Table \ref{tab:limited} and Table \ref{tab:all}. As can be seen in Table \ref{tab:limited}, our model performs comparably as Co-separation with only 0.22 (sigmoid) and 0.19 (softmax) performance degradation in SDR. These results show that our fully unsupervised model is capable of separating sounds at similar quality as Co-separation, even though being validated on categories that were used for pre-training of the object detector in Co-separation.
Moreover, as can be seen in Table \ref{tab:limited} and Table \ref{tab:all}, when used with the softmax constraint, our model performs better than Pixelplayer, the existing fully unsupervised method, by 1.21 (all categories) and 0.66 (limited categories) in SDR. From these results we can deduce that the softmax constraint, which enforces the sum of the separated sound to be equal to the original sound, largely enhances the separation accuracy. It should be noted that this constraint can not be incorporated in Pixelplayer.
\\
\\
\noindent
\textbf{Qualitative evaluation of our model} 
\begin{figure*}[h]
    \scalebox{0.8}{
        \begin{tabular}{p{7.1em} p{7.6em} p{7.6em} p{7.6em} p{7.6em} p{7.6em}}
            \hfil \hfil 
            & \hfil \centering Example 1 \hfil 
            & \hfil \centering Example 2 \hfil 
            & \hfil \centering Example 3 \hfil
            & \hfil \centering Example 4 \hfil 
            & \hfil \centering Example 5 \hfil \tabularnewline

            \begin{minipage}{7.1em}
                \centering
                \scalebox{1}
                {Ours}
            \end{minipage} &
            \begin{minipage}{7.6em}
                \centering
                \scalebox{0.380}
                {\includegraphics{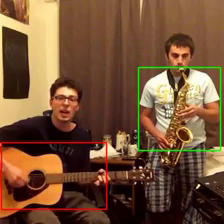}}
            \end{minipage} &
            \begin{minipage}{7.6em}
                \centering
                \scalebox{0.380}
                {\includegraphics{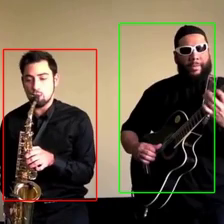}}
            \end{minipage} &
            \begin{minipage}{7.6em}
                \centering
                \scalebox{0.380}
                {\includegraphics{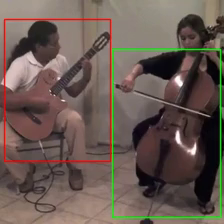}}
            \end{minipage} &
            
            \begin{minipage}{7.6em}
                \centering
                \scalebox{0.380}
                {\includegraphics{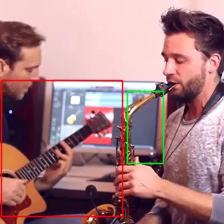}}
            \end{minipage} &
            \begin{minipage}{7.6em}
                \centering
                \scalebox{0.380}
                {\includegraphics{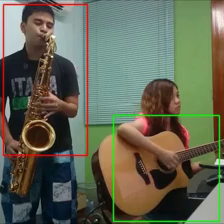}}
            \end{minipage} \vspace{0.001mm} \\
            \begin{minipage}{7.1em}
                \centering
                \scalebox{1}
                {Co-Separation \cite{co_separating}}
            \end{minipage} &
            \begin{minipage}{7.6em}
                \centering
                \scalebox{0.380}
                {\includegraphics{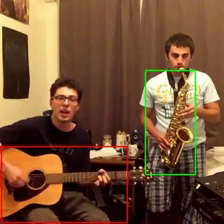}}
            \end{minipage} &
            \begin{minipage}{7.6em}
                \centering
                \scalebox{0.380}
                {\includegraphics{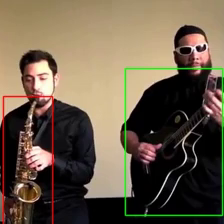}}
            \end{minipage} &
            \begin{minipage}{7.6em}
                \centering
                \scalebox{0.380}
                {\includegraphics{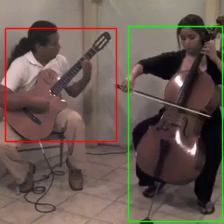}}
            \end{minipage} &
            
            \begin{minipage}{7.6em}
                \centering
                \scalebox{0.380}
                {\includegraphics{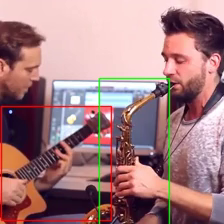}}
            \end{minipage} &
            \begin{minipage}{7.6em}
                \centering
                \scalebox{0.380}
                {\includegraphics{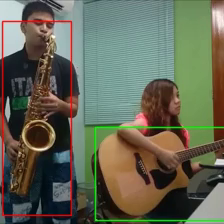}}
            \end{minipage} \vspace{0.001mm} \\
        \end{tabular}     }
    	\caption{The detection results of our model and Co-separation in inference phase. As can be seen in example 2, the detected region sometimes covers the human in addition to the instrument. Similarly, example 1, 4, and 5 show that the regions sometimes do not cover the entire region of the instruments, but instead only covers a specific part. These models are trained on the 7 single instrument categories that are available in the Open Images V4 dataset.}
    \label{fig:duet_comparison}
\end{figure*}
We show examples of the separation and the detection results on the validation set in Fig. \ref{fig:qualitative_solo}. As can be seen in Fig. \ref{fig:qualitative_solo}, our model is applicable to various types of categories, including the ones that are not included in the Open Images V4 Dataset. From the comparison between the predicted and ground truth spectrograms, we can confirm that the separation is properly performed. For the separation and detection results on the test set, please refer to the supplementary video.
\\
\\
\noindent
\textbf{Limitation of our model} 
We show some examples of the detection result of our model and Co-separation in Fig. \ref{fig:duet_comparison}. While Co-separation, which is based on a pre-trained detector, successfully detects the region of the instruments, our model sometimes detects slightly larger or smaller regions. This can be because our model has no explicit constraint that induces the model to detect the exact region. Specifically, if the selected bounding box includes the region that enables the type of object to be identified, the model can reduce the defined loss.

\section{Discussion \& Conclusion}
In this paper, we proposed a fully unsupervised method for audio-visual sound source separation that learns to both detect objects in an image and separate sound sources. While existing methods rely on a pre-trained object detector, our method can be trained in an end-to-end manner including the object detection part. Therefore, unlike the existing methods that rely on a pre-trained detector, our method is applicable to categories whose annotations are not available. Moreover, even though our method does not use any annotations, our method performs comparably in terms of separation accuracy as the method that is based on a pre-trained detector. Additionally, our method performs much better than the existing fully unsupervised method by introducing a simple constraint, which enforces the sum of the separated sound to be equal to the original sound. For future work, we believe it is valuable to try to make audio-visual models applicable to images in which a large number of sound-producing objects are present.

%
\IEEEpeerreviewmaketitle
\textbf{Acknowledgements}
This research was fully supported by JST Mirai Program No. JPMJMI19B2, and JSPS KAKENHI Nos. 19H01129, 19H04137, 21H0504.

\bibliography{egbib}
\bibliographystyle{unsrt}

\end{document}